\documentclass[11pt,letterpaper]{article}
\usepackage[breaklinks=true]{hyperref}
\usepackage{emnlp2017}
\usepackage{times}
\usepackage{latexsym}

\usepackage{amsmath}
\usepackage{color}
\usepackage{comment}
\usepackage{xspace}

\usepackage{tikz}
\usepackage{tikz-qtree}
\usetikzlibrary{positioning}

\newcommand{\open}{\textsc{Open}\xspace}
\newcommand{\close}{\textsc{Close}\xspace}
\newcommand{\shift}{\textsc{Shift}\xspace}

\makeatletter
\DeclareRobustCommand{\rvdots}{%
  \vbox{
    \baselineskip4\p@\lineskiplimit\z@
    \kern-\p@
    \hbox{.}\hbox{.}\hbox{.}
  }}
\makeatother

\emnlpfinalcopy

\def\emnlppaperid{***}

\title{Effective Inference for Generative Neural Parsing}

\author{
  Mitchell Stern \qquad Daniel Fried \qquad Dan Klein \\
  Computer Science Division \\
  University of California, Berkeley \\
  {\tt \{mitchell,dfried,klein\}@cs.berkeley.edu}
}

\date{}

\begin{document}

\maketitle

\begin{abstract}
Generative neural models have recently achieved state-of-the-art results for constituency parsing. However, without a feasible search procedure, their use has so far been limited to reranking the output of external parsers in which decoding is more tractable. We describe an alternative to the conventional action-level beam search used for discriminative neural models that enables us to decode directly in these generative models. We then show that by improving our basic candidate selection strategy and using a coarse pruning function, we can improve accuracy while exploring significantly less of the search space. Applied to the model of \citet{Choe16Parsing}, our inference procedure obtains 92.56 F1 on section 23 of the Penn Treebank, surpassing prior state-of-the-art results for single-model systems.
\end{abstract}

\section{Introduction}

A recent line of work has demonstrated the success of generative neural models for constituency parsing \cite{dyer2016recurrent,Choe16Parsing}. As with discriminative neural parsers, these models lack a dynamic program for exact inference due to their modeling of unbounded dependencies. However, while discriminative neural parsers are able to obtain strong results using greedy search \cite{dyer2016recurrent} or beam search with a small beam \cite{vinyals2015grammar}, we find that a simple action-level approach fails outright in the generative setting. Perhaps because of this, the application of generative neural models has so far been restricted to reranking the output of external parsers.

Intuitively, because a generative parser defines a joint distribution over sentences and parse trees, probability mass will be allocated unevenly between a small number of common structural actions and a large vocabulary of lexical items. This imbalance is a primary cause of failure for search procedures in which these two types of actions compete directly. A notion of equal competition among hypotheses is then desirable, an idea that has previously been explored in generative models for constituency parsing \cite{henderson2003inducing} and dependency parsing \cite{titov2010latent,buys2015generative}, among other tasks. We describe a related state-augmented beam search for neural generative constituency parsers in which lexical actions compete only with each other rather than with structural actions. Applying this inference procedure to the generative model of \citet{Choe16Parsing}, we find that it yields a self-contained generative parser that achieves high performance.

Beyond this, we propose an enhanced candidate selection strategy that yields significant improvements for all beam sizes. Additionally, motivated by the look-ahead heuristic used in the top-down parsers of \citet{roark2001probabilistic} and \citet{charniak2010top}, we also experiment with a simple coarse pruning function that allows us to reduce the number of states expanded per candidate by several times without compromising accuracy. Using our final search procedure, we surpass prior state-of-the-art results among single-model parsers on the Penn Treebank, obtaining an F1 score of 92.56.

\section{Common Framework}
\label{sec:common-framework}

\begin{figure}
\centering
\begin{tikzpicture}[level distance=25pt]
\Tree [. S [. NP He ] [. VP had [. NP an idea ] ] {.} ]
\node[inner sep=0] at (0, -3.5) {(S (NP He NP) (VP had (NP an idea NP) VP) . S)};
\end{tikzpicture}
\caption{A parse tree and the action sequence that produced it, corresponding to the sentence ``He had an idea.'' The tree is constructed in left-to-right depth-first order. The tree contains only nonterminals and words; part-of-speech tags are not included. $\open(X)$ and $\close(X)$ are rendered as ``$(X$'' and ``$X)$'' for brevity.}
\label{fig:example-tree}
\end{figure}
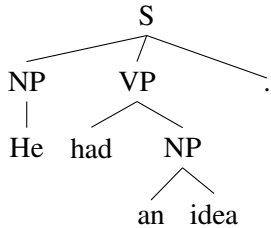

The generative neural parsers of \citet{dyer2016recurrent} and \citet{Choe16Parsing} can be unified under a common shift-reduce framework. Both systems build parse trees in left-to-right depth-first order by
executing a sequence of actions, as illustrated in Figure~\ref{fig:example-tree}. These actions can be grouped into three major types:
$\open(X)$ and $\close(X)$, which open and close a constituent with nonterminal $X$,\footnote{The model described in \citet{dyer2016recurrent} has only a single \close action, whereas the model described in \citet{Choe16Parsing} annotates $\close(X)$ actions with their nonterminals. We present the more general version here.} respectively, and $\shift(x)$, which adds the word $x$ to the current constituent. The probability of an action sequence $(a_1, \dots, a_T)$ is 
\begin{align*}
P(a_1, \dots, a_T)
  &= \prod_{t=1}^T P(a_t \mid a_1, \dots, a_{t-1}) \\
  &= \prod_{t=1}^T \,\, [\mathrm{softmax}(\mathbf{W} \mathbf{u}_t + \mathbf{b})]_{a_t} ,
\end{align*}
where $\mathbf{u}_t$ is a continuous representation of the parser's state at time $t$, and $[\mathbf{v}]_j$ denotes the $j$th component of a vector $\mathbf{v}$. We refer readers to the respective authors' papers for the parameterization of $\mathbf{u}_t$ in each model.

In both cases, the decoding process reduces to a search for the most probable action sequence that represents a valid tree over the input sentence. For a given hypothesis, this requirement implies several constraints on the successor set \cite{dyer2016recurrent}; e.g., $\shift(x)$ can only be executed if the next word in the sentence is $x$, and $\close(X)$ cannot be executed directly after $\open(X)$.

\section{Model and Training Setup}

We reimplemented the generative model described in \citet{Choe16Parsing} and trained it on the Penn Treebank \cite{Marcus93Building} using their published hyperparameters and preprocessing. However, rather than selecting the final model based on reranking performance, we instead perform early stopping based on development set perplexity. We use sections 2-21 of the Penn Treebank for training, section 22 for development, and section 23 for testing. The model's action space consists of 26 matching pairs of \open and \close actions, one for each nonterminal, and 6,870 \shift actions, one for each preprocessed word type. While we use this particular model for our experiments, we note that our subsequent discussion of inference techniques is equally applicable to any generative parser that adheres to the framework described above in Section~\ref{sec:common-framework}.

\section{Action-Level Search}

\begin{figure}
\centering
\begin{tikzpicture}
\pgfmathsetmacro{\xScale}{0.58}
\pgfmathsetmacro{\barOffsetY}{0.2}
\pgfmathsetmacro{\barScaleY}{0.75}
\pgfmathsetmacro{\barWidth}{0.7}
\pgfmathsetmacro{\textOffsetY}{0.2}
\pgfmathsetmacro{\direction}{-1}
\foreach \word [count=\i] in {{(S},{(NP},He,{NP)},{(VP},{had},{(NP},an,idea,{NP)},{VP)},.,{S)}}
  \node[anchor=base] at (\xScale*\i, 0) {\small\word};
\foreach \prob [count=\i] in {0.1,0.5,3.2,0.0,0.1,4.6,1.4,3.9,4.0,0.1,0.8,0.0,0.0} {
  \draw[draw=black,fill=blue]
    ({\xScale*(\i-\barWidth/2)},\direction*\barOffsetY) rectangle
    ({\xScale*(\i+\barWidth/2)},{\direction*(\barOffsetY+\barScaleY*\prob)});
  \node at (\xScale*\i,{\direction*(\barOffsetY+\barScaleY*\prob+\textOffsetY}) {\small-\prob};
}
\end{tikzpicture}
\caption{A plot of the action log probabilities $\log P(a_t \mid a_1, \dots, a_{t-1})$ for the example in Figure~\ref{fig:example-tree} under our main model. We observe that \open and \close actions have much higher probability than \shift actions. This imbalance is responsible for the failure of standard action-level beam search.}
\label{fig:example-log-probs}
\end{figure}
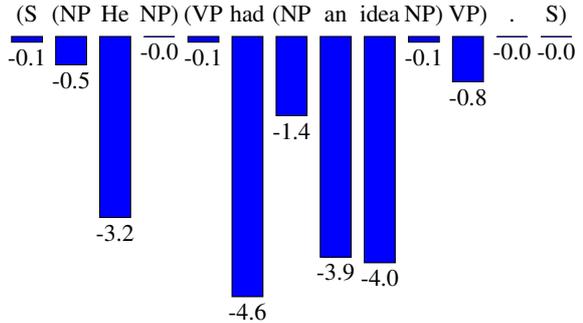

Given that ordinary action-level search has been applied successfully to discriminative neural parsers \cite{vinyals2015grammar,dyer2016recurrent}, it offers a sensible starting point for decoding in generative models. However, even for large beam sizes, the following pathological behavior is encountered for generative decoding, preventing reasonable parses from being found. Regardless of the sequence of actions taken so far, the generative model tends to assign much higher probabilities to structural \open and \close actions than it does lexical \shift actions, as shown in Figure~\ref{fig:example-log-probs}. The model therefore prefers to continually open new constituents until a hard limit is reached, as the alternative at each step is to take the low-probability action of shifting the next word. The resulting sequence typically has much lower overall probability than a plausible parse, but the model's myopic comparison between structural and lexical actions prevents reasonable candidates from staying on the beam. Action-level beam search with beam size 1000 obtains an F1 score of just 52.97 on the development set.

\section{Word-Level Search}
\label{sec:word-level-search}

The imbalance between the probabilities of structural and lexical actions suggests that the two kinds of actions should not compete against each other within a beam. This leads us to consider an augmented state space in which they are kept separate by design, as was done by \citet{Fried17Improving}. In conventional action-level beam search, hypotheses are grouped by the length of their action history $|A|$. Letting $A_i$ denote the set of actions taken since the $i$th shift action, we instead group hypotheses by the pair $(i, |A_i|)$, where $i$ ranges between 0 and the length of the sentence.

Let $k$ denote the target beam size. The search process begins with the empty hypothesis in the $(0, 0)$ bucket. Word-level steps are then taken according to the following procedure for $i = 0, 1, \dots$, up to the length of the sentence (inclusive). Beginning with the $(i, 0)$ bucket, the successors of each hypothesis are pooled together, sorted by score, and filtered down to the top $k$. Of those that remain, successors obtained by taking an \open or \close action advance to the $(i, 1)$ bucket, whereas successors obtained from a \shift action are placed in the $(i+1, 0)$ bucket if $i$ is less than the sentence length, or the completed list if $i$ is equal to the sentence length. This process is repeated for the $(i, 1)$ bucket, the $(i, 2)$ bucket, and so forth, until the $(i+1, 0)$ bucket contains at least $k$ hypotheses. If desired, a separate word beam size $k_w < k$ can be used at word boundaries, in which case each word-level step terminates when the $(i+1, 0)$ bucket has $k_w$ candidates instead of $k$. This introduces a bottleneck that can help to promote beam diversity.

\begin{table}
\centering
\resizebox{\linewidth}{!}{\begin{tabular}{c|cccccc}
Beam Size $k$ & 200 & 400 & 600 & 800 & 1000 & 2000 \\ \hline
$k_w = k$ & 87.47 & 89.86 & 90.98 & 91.62 & 91.97 & 92.74 \\
$k_w = k/10$ & 89.25 & 91.16 & 91.83 & 92.12 & 92.38 & 92.93 \\
\end{tabular}}
\caption{Development F1 scores using word-level search with various beam sizes $k$ and two choices of word beam size $k_w$.}
\label{tab:word-level-results}
\end{table}

Development set results for word-level search with a variety of beam sizes and with $k_w = k$ or $k_w = k/10$ are given in Table~\ref{tab:word-level-results}. We observe that performance in both cases increases steadily with beam size. Word-level search with $k_w = k/10$ consistently outperforms search without a bottleneck at all beam sizes, indicating the utility of this simple diversity-inducing modification. The top result of 92.93 F1 is already quite strong compared to other single-model systems.

\section{Fast-Track Candidate Selection}
\label{sec:fast-track}

\begin{figure}
\centering
\begin{tikzpicture}[xscale=1.9,yscale=0.7]

\tikzset{
  >=latex,
  minimum height=1.5em,
  minimum width=2.5em,
  normal/.style={draw},
  top-k/.style={draw,fill=blue!10},
  fast-track/.style={draw,fill=red!20},
}

\node[normal] (1-1) at (0,4) {had};
\node[normal] (1-2) at (0,0) {had};
\node[normal] (1-3) at (0,-4) {had};

\draw[->,dashed] (1-1) +(-0.5,0) -- (1-1);
\draw[->,dashed] (1-2) +(-0.5,0) -- (1-2);
\draw[->,dashed] (1-3) +(-0.5,0) -- (1-3);

\begin{scope}[shift={(1,2.5)}]
\node[top-k] (2-1-1) at (0,3) {(NP};
\node[normal] (2-1-2) at (0,2) {(PP};
\node (2-1-3) at (0,1) {\rvdots};
\node[normal] (2-1-4) at (0,0) {an};
\end{scope}

\begin{scope}[shift={(1,-1.5)}]
\node[top-k] (2-2-1) at (0,3) {(S};
\node[normal] (2-2-2) at (0,2) {(NP};
\node (2-2-3) at (0,1) {\rvdots};
\node[fast-track] (2-2-4) at (0,0) {an};
\end{scope}

\begin{scope}[shift={(1,-5.5)}]
\node[top-k] (2-3-1) at (0,3) {(NP};
\node[normal] (2-3-2) at (0,2) {(S};
\node (2-3-3) at (0,1) {\rvdots};
\node[normal] (2-3-4) at (0,0) {an};
\end{scope}

\path[->]
  (1-1.east) edge (2-1-1.west) edge (2-1-2.west) edge (2-1-3.west) edge (2-1-4.west)
  (1-2.east) edge (2-2-1.west) edge (2-2-2.west) edge (2-2-3.west) edge (2-2-4.west)
  (1-3.east) edge (2-3-1.west) edge (2-3-2.west) edge (2-3-3.west) edge (2-3-4.west)
;

\begin{scope}[shift={(2,2.5)}]
\node[top-k] (3-1-1) at (0,3) {(NP};
\node[normal] (3-1-2) at (0,2) {(VP};
\node (3-1-3) at (0,1) {\rvdots};
\node[fast-track] (3-1-4) at (0,0) {an};
\end{scope}

\begin{scope}[shift={(2,-1.5)}]
\node[normal] (3-2-1) at (0,3) {(NP};
\node[normal] (3-2-2) at (0,2) {(NX};
\node (3-2-3) at (0,1) {\rvdots};
\node[normal] (3-2-4) at (0,0) {an};
\end{scope}

\begin{scope}[shift={(2,-5.5)}]
\node[top-k] (3-3-1) at (0,3) {an};
\node[top-k] (3-3-2) at (0,2) {(NP};
\node (3-3-3) at (0,1) {\rvdots};
\node[normal] (3-3-4) at (0,0) {(QP};
\end{scope}

\path[->]
  (2-1-1.east) edge (3-1-1.west) edge (3-1-2.west) edge (3-1-3.west) edge (3-1-4.west)
  (2-2-1.east) edge (3-2-1.west) edge (3-2-2.west) edge (3-2-3.west) edge (3-2-4.west)
  (2-3-1.east) edge (3-3-1.west) edge (3-3-2.west) edge (3-3-3.west) edge (3-3-4.west)
;

\begin{scope}[shift={(3.25,0)}]
\node[normal] (4-1) at (0,4) {an};
\node[normal] (4-2) at (0,0) {an};
\node[normal] (4-3) at (0,-4) {an};
\end{scope}

\node (1-caption) at (0,7) {$(i,0)$};
\node (2-caption) at (1,7) {$(i,1)$};
\node (3-caption) at (2,7) {$(i,2)$};
\node (4-caption) at (3.25,7) {$(i+1,0)$};

\node (2-above-right) at (1.5,6.25) {};
\node (3-above-right) at (2.5,6.25) {};
\draw[->,dashed] (2-2-4.east) -| (2-above-right.center) -- (3-above-right.center) |- (4-1.west);
\draw[->,dashed] (3-1-4.east) -- (3-1-4.east -| 3-above-right.center) |- (4-2.west);
\draw[->,dashed] (3-3-1.east) -- (3-3-1.east -| 3-above-right.center) |- (4-3.west);

\node (separator) at (2.75,0) {};
\draw[dotted] (separator |- 4-caption.north) -- (separator |- 3-3-4.south);

\end{tikzpicture}\\[1em]
\begin{tikzpicture}
\tikzset{minimum width=1em,minimum height=1em}
\node[rectangle,fill=blue!10] (top-k-color) {\phantom{x}};
\node[base right=0 of top-k-color] (top-k-text) {top-$k$};
\node[base right=2em of top-k-text,rectangle,fill=red!20] (fast-track-color) {\phantom{x}};
\node[base right=0 of fast-track-color] (fast-track-text) {fast-track};
\end{tikzpicture}
\caption{One step of word-level search with fast-track candidate selection (Sections~\ref{sec:word-level-search} and~\ref{sec:fast-track}) for the example in Figure~\ref{fig:example-tree}. Grouping candidates by the current word $i$ ensures that low-probability lexical actions are kept separate from high-probability structural actions at the beam level. Fast-track selection mitigates competition between the two types of actions within a single pool of successors.}
\label{fig:beam-search}
\end{figure}
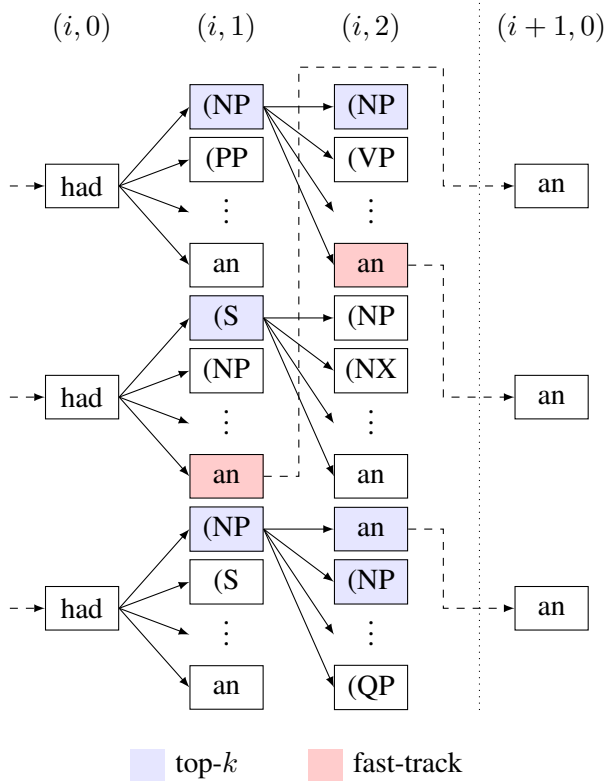

The word-level beam search described in Section~\ref{sec:word-level-search} goes one step toward ameliorating the issue that causes action-level beam search to fail, namely the direct competition between common structural actions with high probabilities and low-frequency shift actions with low probabilities. However, the issue is still present to some extent, in that successors of both types from a given bucket are pooled together and filtered down as a single collection before being routed to their respective destinations. We therefore propose a more direct solution to the problem, in which a small number $k_s \ll k$ of \shift successors are fast-tracked to the next word-level bucket \emph{before} any filtering takes place. These fast-tracked candidates completely bypass competition with potentially high-scoring \open or \close successors, allowing for higher-quality results in practice with minimal overhead. See Figure~\ref{fig:beam-search} for an illustration.

\begin{table}
\centering
\resizebox{\linewidth}{!}{\begin{tabular}{c|cccccc}
Beam Size $k$ & 200 & 400 & 600 & 800 & 1000 & 2000 \\ \hline
$k_w = k$ & 91.33 & 92.17 & 92.51 & 92.73 & 92.89 & 93.05 \\
$k_w = k/10$ & 91.41 & 92.34 & 92.70 & 92.94 & 93.09 & 93.18 \\
\end{tabular}}
\caption{Development F1 scores using the settings from Table~\ref{tab:word-level-results}, together with the fast-track selection strategy from Section~\ref{sec:fast-track} with $k_s = k/100$.}
\label{tab:fast-track-results}
\end{table}

We repeat the experiments from Section~\ref{sec:word-level-search} with $k_s = k/100$ and report the results in Table~\ref{tab:fast-track-results}. Note that the use of fast-tracked candidates offers significant gains under all settings.
The top result improves from 92.93 to 93.18 with the use of fast-tracked candidates, surpassing prior single-model systems on the development set.

\section{\open Action Pruning}

At any point during the trajectory of a hypothesis, either 0 or all 26 of the \open actions will be available, compared with at most 1 \close action and at most 1 \shift action. Hence, when available, \open actions comprise most or all of a candidate's successor actions. To help cut down on this portion of the search space, it is natural to consider whether some of these actions could be ruled out using a coarse model for pruning.

\subsection{Coarse Model}

We consider a class of simple pruning models that condition on the $c \ge 0$ most recent actions and the next word in the sentence, and predict a probability distribution over the next action. In the interest of efficiency, we collapse all \shift actions into a single unlexicalized \shift action, significantly reducing the size of the output vocabulary.

The input $\mathbf{v}_t$ to the pruning model at time $t$ is the concatenation of a vector embedding for each action in the context $(a_{t-c}, a_{t-c+1}, \dots, a_{t-1})$ and a vector embedding for the next word $w$:
\begin{align*}
\mathbf{v}_t = [\mathbf{e}_{a_{t-c}}; \mathbf{e}_{a_{t-c+1}}; \dots; \mathbf{e}_{a_{t-1}}; \mathbf{e}_{w}],
\end{align*}
where each $\mathbf{e}_j$ is a learned vector embedding.
The pruning model itself is implemented by feeding the input vector through a one-layer feedforward network with a ReLU non-linearity, then applying a softmax layer on top:
\begin{align*}
& P(a_t = a \mid a_1, \dots, a_{t-1}, \text{next-word} = w) \\
  &= P(a_t = a \mid a_{t-c}, \dots, a_{t-1}, \text{next-word} = w) \\
  &= [\mathrm{softmax}(\mathbf{W}_2 \max(\mathbf{W}_1 \mathbf{v}_t + \mathbf{b}_1, 0) + \mathbf{b}_2)]_a .
\end{align*}
The pruning model is trained separately from the main parsing model on gold action sequences derived from the training corpus, with log-likelihood as the objective function and a cross entropy loss.

\subsection{Strategy and Empirical Lower Bound}

Once equipped with a coarse model, we use it for search reduction in the following manner. As mentioned above, when a hypothesis is eligible to open a new constituent, most of its successors will be obtained through \open actions. Accordingly, we use the coarse model to restrict the set of \open actions to be explored. When evaluating the pool of successors for a given collection of hypotheses during beam search, we run the coarse model on each hypothesis to obtain a distribution over its next possible actions, and gather together all the coarse scores of the would-be \open successors. We then discard the \open successors whose coarse scores lie below the top $1 - p$ quantile for a fixed $0 < p < 1$, guaranteeing that no more than a $p$-fraction of \open successors are considered for evaluation. Taking $p = 1$ corresponds to the unpruned setting.

This strategy gives us a tunable hyperparameter $p$ that allows us to trade off between the amount of search we perform and the quality of our results. Before testing our procedure, however, we would first like to investigate whether there is a principled bound on how low we can expect to set $p$ without a large drop in performance. A simple estimate arises from noting that the pruning fraction $p$ should be set to a value for which most or all of the outputs encountered in the training set are retained. Otherwise, the pruning model would prevent the main model from even recreating the training data, let alone producing good parses for new sentences.

\begin{table}
\centering
\resizebox{\linewidth}{!}{\begin{tabular}{c||cccccccccc}
$c$ & 1 & 2 & 3 & 4 & 5 & 6 & 7 & 8 & 9 & 10 \\ \hline
0 & 20.0 & 58.4 & 82.4 & 91.0 & 94.9 & 96.8 & 97.9 & 98.6 & 98.9 & 99.2 \\
1 & 54.9 & 80.5 & 91.1 & 95.9 & 97.7 & 98.8 & 99.5 & 99.8 & 99.9 & 100.0 \\
2 & 61.2 & 85.0 & 93.8 & 97.4 & 98.6 & 99.5 & 99.8 & 99.9 & 100.0 & 100.0 \\
\end{tabular}}
\caption{Cumulative distributions of the number of unique \open outputs per input for an order-$c$ pruning function, computed over pruning inputs with at least one \open output.}
\label{tab:prune-cumulative-distributions}
\end{table}

To this end, we collect training corpus statistics on the occurrences of inputs to the pruning function and their corresponding outputs.
We then compute the number of unique \open actions associated with inputs occurring at least 20 times, and restrict our attention to inputs with at least one \open output. The resulting cumulative distributions for context sizes $c = 0, 1, 2$ are given in Table~\ref{tab:prune-cumulative-distributions}. If we require that our pruning fraction $p$ be large enough to recreate at least 99\% of the training data, then since there are 26 total nonterminals, approximate\footnote{These thresholds are not exact due to the fact that our pruning procedure operates on collections of multiple hypotheses' successors at inference time rather than the successors of an individual hypothesis.} lower bounds for $p$ are
$10/26 \approx 0.385$ for $c = 0$, $7/26 \approx 0.269$ for $c = 1$, and $6/26 \approx 0.231$ for $c = 2$.

\subsection{Pruning Results}

\begin{table}
\resizebox{\linewidth}{!}{\begin{tabular}{c|ccccccc}
$p$ & 6/26 & 7/26 & 8/26 & 9/26 & 10/26 & 11/26 & 1 \\ \hline
Dev F1 & 92.78 & 93.00 & 93.08 & 93.13 & 93.19 & 93.19 & 93.18 \\
\end{tabular}}
\caption{Results when the best setting from Section~\ref{sec:fast-track} is rerun with \open action pruning with context size $c = 2$ and various pruning fractions $p$. Lower values of $p$ indicate more aggressive pruning, while $p = 1$ means no pruning is performed.}
\label{tab:results-pruning}
\end{table}

We reran our best experiment from Section~\ref{sec:fast-track} with an order-2 pruning function and pruning fractions $p = 6/26, \dots, 11/26$. The results are given in Table~\ref{tab:results-pruning}. We observe that performance is on par with the unpruned setup (at most 0.1 absolute difference in F1 score) for $p$ as low as $8/26 \approx 0.308$. Setting $p$ to $7/26 \approx 0.269$ results in a drop of 0.18, and setting $p$ to $6/26 \approx 0.231$ results in a drop of 0.40. Hence, degradation begins to occur right around the empirically-motivated threshold of $6/26$ given above, but we can prune $1 - 8/26 \approx 69.2\%$ of \open successors with minimal changes in performance.

\section{Final Results and Conclusion}

\begin{table}
\resizebox{\linewidth}{!}{\begin{tabular}{l|ccc}
Parser & LR & LP & F1 \\ \hline
\citet{vinyals2015grammar} & -- & -- & 88.3 \\
\citet{Shindo12Bayesian} & -- & -- & 91.1 \\
\citet{Cross16Span} & 90.5 & 92.1 & 91.3 \\
\citet{dyer2016recurrent} & -- & -- & 91.7 \\
\citet{Liu17ShiftReduce} & 91.3 & 92.1 & 91.7 \\
\citet{Stern17Minimal} & 90.63 & 92.98 & 91.79 \\ \hline
Our Best Result & 92.57 & 92.56 & 92.56 \\
Our Best Result (with pruning) & 92.52 & 92.54 & 92.53 \\ \hline
\citet{vinyals2015grammar} (ensemble) & -- & -- & 90.5 \\
\citet{Shindo12Bayesian} (ensemble) & -- & -- & 92.4 \\
\citet{Choe16Parsing} (rerank) & -- & -- & 92.6 \\
\citet{dyer2016recurrent} (rerank) & -- & -- & 93.3 \\
\citet{Fried17Improving} (ensemble, rerank) & -- & -- & 94.25 \\
\end{tabular}}
\caption{Comparison of F1 scores on section 23 of the Penn Treebank. Here we only include models trained without external silver training data. Results in the first two sections are for single-model systems.}
\label{tab:final-results}
\end{table}

We find that the best overall settings are a beam size of $k = 2000$, a word beam size of $k_w = 200$, and $k_s = 20$ fast-track candidates per step, as this setup achieves both the highest probabilities under the model and the highest development F1. We report our test results on section 23 of the Penn Treebank under these settings in Table~\ref{tab:final-results} both with and without pruning, as well as a number of other recent results. We achieve F1 scores of 92.56 on the test set without pruning and 92.53 when $1 - 8/26 \approx 69.2\%$ of \open successors are pruned, obtaining performance well above the previous state-of-the-art scores for single-model parsers. This demonstrates that the model of \citet{Choe16Parsing} works well as an accurate, self-contained system. The fact that we match the performance of their reranking parser using the same generative model confirms the efficacy of our approach. We believe that further refinements of our search procedure can continue to push the bar higher, such as the use of a learned heuristic function for forward score estimation, or a more sophisticated approximate decoding scheme making use of specific properties of the model. We look forward to exploring these directions in future work.

\section*{Acknowledgments}

MS is supported by an NSF Graduate Research Fellowship. DF is supported by an NDSEG fellowship.

\bibliography{paper}
\bibliographystyle{emnlp_natbib}

\end{document}